\pdfoutput=1

\documentclass[11pt]{article}
\usepackage[table]{xcolor}
\usepackage{EMNLP2022}
\usepackage{times}
\usepackage{latexsym}

\usepackage[T1]{fontenc}

\usepackage[utf8]{inputenc}
\usepackage{microtype}
\usepackage[most]{tcolorbox}
\usepackage{algorithm}
\usepackage{anyfontsize}
\usepackage{dirtytalk}
\usepackage{geometry}
\usepackage{booktabs}
\usepackage{amsmath}
\usepackage[normalem]{ulem}
\useunder{\uline}{\ul}{}
\usepackage{booktabs}  
\usepackage{graphicx} 
\usepackage{listings}
\usepackage[wby]{callouts}
\usepackage{t1enc}
\usepackage{booktabs}
\usepackage{times}
\usepackage{multirow}
\usepackage{graphicx}
\usepackage{amsmath}
\usepackage{algorithm}
\usepackage[noend]{algpseudocode}
\usepackage{diagbox}
\usepackage{latexsym}
\usepackage{enumitem}
\usepackage{siunitx}

\usepackage{helvet}  
\usepackage{courier}  
\usepackage{url}  
\usepackage{graphicx}  
\usepackage{courier}
\usepackage{amsmath}
\usepackage[normalem]{ulem}
\useunder{\uline}{\ul}{}
\usepackage{graphicx}
\usepackage{multirow}
\usepackage{mathbbol}
\usepackage{caption}
\usepackage{CJKutf8}
\usepackage{verbatim}
\usepackage{float}
\usepackage{booktabs}
\usepackage{array}
\usepackage{times}
\usepackage{csquotes}
\usepackage{latexsym}
\usepackage{tablefootnote}
\usepackage[normalem]{ulem}
\usepackage{todonotes}
\usepackage{subcaption}
\usepackage{titlesec}
\usepackage{tikz}
\usepackage{bbding}
\usepackage{pifont}
\usepackage{wasysym}
\usepackage{amssymb}

\usepackage{microtype}

\titlespacing*{\section}
{0pt}{.5ex plus .5ex minus .2ex}{.5ex plus .2ex minus .2ex}
\titlespacing*{\subsection}
{0pt}{.5ex plus .5ex minus .2ex}{.5ex plus .2ex minus .2ex}

\title{A Unified Framework for Pun Generation with Humor Principles}
\author{
Yufei Tian\textsuperscript{1},
Divyanshu Sheth\textsuperscript{2}\thanks{~~Work done when the author was interning at UCLA.},
  Nanyun Peng\textsuperscript{1} \\
 \textsuperscript{1}Computer Science Department, University of California, Los Angeles, \\
   \textsuperscript{2}Department of Industrial and Systems Engineering, Indian Institute of Technology, Kharagpur\\
  \texttt{ \{yufeit, violetpeng\}@cs.ucla.edu}, \texttt{divyanshusheth1@gmail.com} \\
}

\begin{document}
\maketitle
\begin{abstract}
We propose a unified framework to generate both homophonic and homographic puns to resolve the split-up in existing works. Our framework takes a theoretically motivated approach to incorporate three linguistic attributes of puns to language models: ambiguity, distinctiveness, and surprise. Our framework consists of three parts: 1) a context words/phrases selector to promote the aforementioned humor attributes, 2) a generation model trained on non-pun sentences to incorporate the context words/phrases into the generation output, and 3) a label predictor that learns the \textit{structure} of puns which is used to steer the generation model at inference time. 
Evaluation results on both homophonic and homographic puns demonstrate the superiority of our model over strong baselines.\footnote{Our code is available at \url{https://github.com/PlusLabNLP/Unified_PunGen}}
\end{abstract}

\section{Introduction}

\begin{figure}[t!]
  \centering
    \includegraphics[width=0.45\textwidth]{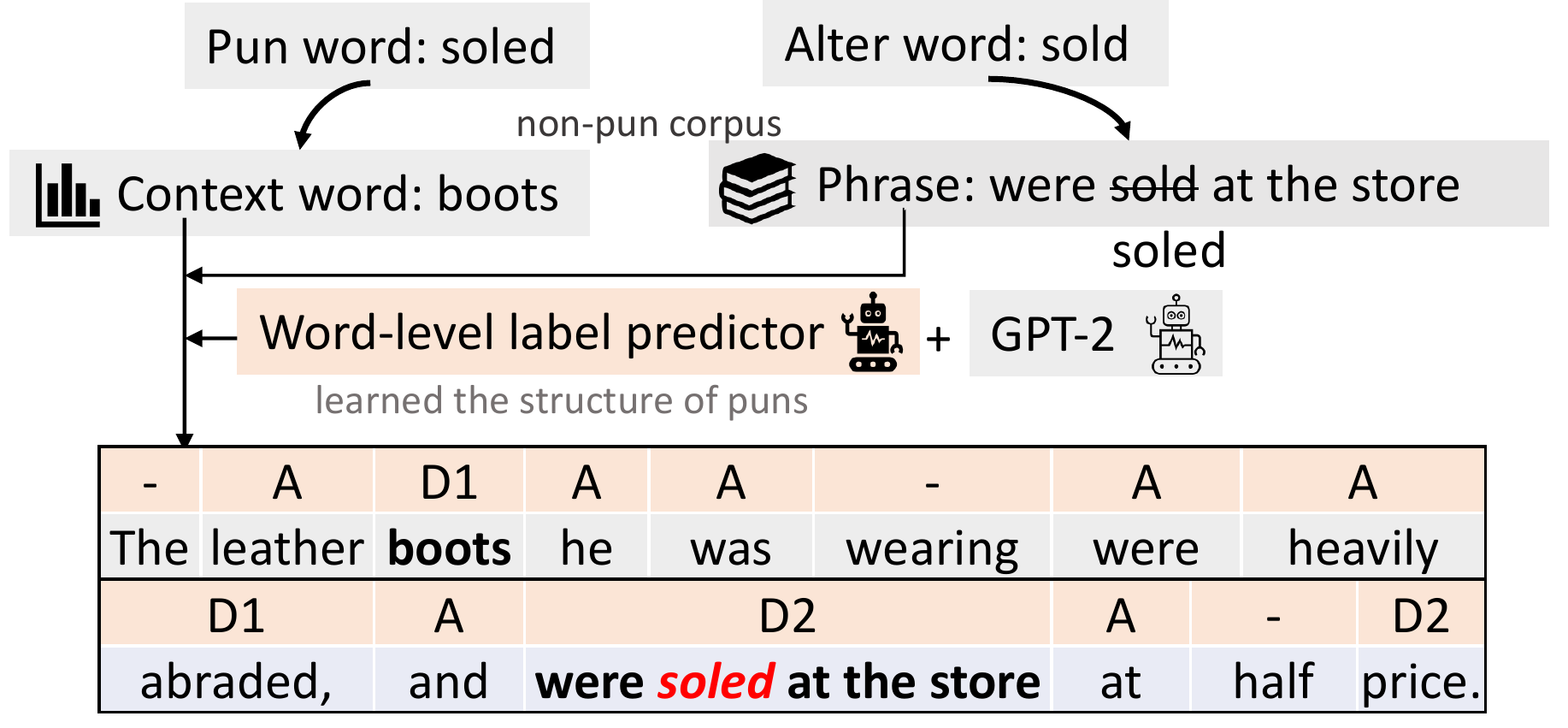}
  \caption{\footnotesize An illustration of our approach. The pun word pair (e.g. \textit{`soled-sold'}) is the input. After retrieving a suitable context word and a phrase, we use a pun label predictor to steer the base GPT-2 model to produce puns. Labels D1/D2/A mean the next word should be distinct to (supporting) the pun word, distinct to (supporting) the alternative word, or ambiguous. A `-' mark means the label predictor is less confident and thus we do not intervene the generation process.
  }
  \label{fig:illustration}    
  \vspace{-5.7 mm}
\end{figure}

Recently, computational humor theories investigating why puns are funny have shown high correlations with human judgments. \citet{kao2016computational} use a probabilistic model to decompose puns into two dimensions: \textbf{ambiguity} of meaning and \textbf{distinctiveness} of viewpoints, and show that these two aspects combined have the strongest alignment with human judgments (p<5\textperthousand).  \citet{he2019pun} show that ambiguity/distincitveness alone cannot capture the whole picture, and 
develop an additional metric to measure how much \textbf{surprise} is aroused when the pun word and alternative word are flipped. For example in Figure \ref{fig:illustration}, the pun word is \textit{soled} and the alternative word is \textit{sold}. Seeing \textit{soled} in the phrase `were \textit{soled} at the store at half price' instead of \textit{sold} arouses surprise in the local context but makes sense in the global context.

Despite the success in identifying important linguistic traits of successful pun sentences, how to incorporate these aforementioned theories into the pun generation process is still an open problem.  Although \citet{he2019pun} propose a retrieve-and-edit approach to incorporate surprise, their error analysis shows that the proposed retrieval methods are often unsuccessful.

Moreover, existing works on pun generation are split up in terms of generating \textbf{homographic puns}, wherein the same written word has two or more meanings \cite{mittal2022ambipun, yu-etal-2020-homophonic, yu2018neural}, and \textbf{homophonic puns}, where two words that sound similar have different meanings \cite{luo2019pun, he2019pun, hashimoto2018retrieveandedit}. There lacks a unified generation framework for both types of puns. 

In this work, we incorporate all three principles: ambiguity, distinctiveness, and surprise into pun generation, and bridge the gap between the two pun types. We hypothesize that \textit{there is a \textbf{learnable structure} for puns regardless of the pun type}, and propose a unified framework by converting homographic puns to homophonic ones. Specifically, we carefully extract from a non-pun corpus 1) a context word that supports the meaning of the pun word, and 2) a phrase that is both characteristic to the alternative word and compatible with the pun word. Next, we train a discriminator on existing homophonic pun data to learn the structure of a pun -- the \textit{type} of each word in the sentence, which could be one of `A' -- ambiguous, `D1' -- distinct to the pun word, or `D2' -- distinct to the alternative word. One challenge, however, is that there are no ground truth labels. To this end, we collect a small amount of human annotations and boost from weak, unsupervised models to stronger, supervised models. At inference time, a label predictor is used to guide a base GPT-2 model to generate puns. At each generation step, we re-score the tokens generated by the base language model according to the predicted type, except for the case when the label predictor's confidence is under a set threshold. Our model outperforms existing baselines for both pun types.

\section{Related Works}
\paragraph{Linguistic traits of puns.} \citet{kao2016computational} decompose puns into two dimensions — \textit{ambiguity} of meaning and \textit{distinctiveness} of viewpoints, and show that that ambiguity is useful in distinguishing non-puns from puns, while distinctiveness is useful when spotting good and funny puns from bad or boring non-puns. To the best of our knowledge, we are the first to formally incorporate the famous ambiguity-distinctiveness principle to guide pun generation. In addition, \citet{he2019pun} propose the \textit{local-global surprisal principle} to measure the humorous effect aroused when a word appears unexpectedly in the local context but makes sense given the global context, based on which we improve the way surprise is introduced in generation.

\paragraph{Pun generation.}
 Existing works on pun generation often rely on naive intuitions of semantic ambivalence. For example, \citet{yu2018neural} and \citet{luo2019pun} promote the ambivalence of the pun word via a constrained language model and reinforcement learning; others find related words to support semantic ambiguity \cite{yu-etal-2020-homophonic,mittal2022ambipun}. However, these systems lack serious theoretical backbones and therefore none could evaluate their generated results with regards to the proposed intuitions. What's more, the nature of `ambivalence' alone leads to generic/boring word choice and short outputs. By incorporating distinctiveness and surprise, we ensure that the generated puns are informative and interesting.
 
One reason that previous works leverage those simple intuitions to generate puns \cite{he2019pun, yu-wan-2019-avoid, yu-etal-2020-homophonic} is that  the small corpora size \cite{miller2017semeval, sun2022expun} makes it impractical to train generation models end-to-end using human written puns. We hence propose to learn the \textit{structure} of puns instead of the \textit{actual texts}, which requires far less data to train on. Finally, all previous works (except a concurrent one \cite{sun2022contextpun}) can only generate either homographic puns or homophonic puns. Leveraging the shared structure of puns regardless of the pun type, our model can generate both pun types.


\section{Methodology}\label{sec:method}
The input to our system is a pun word-alternative word pair ($pw$-$aw$, e.g., \textit{soled-sold}), and the target output is a high-quality pun sentence that contains $pw$, e.g., \textit{`The leather boots he was wearing were heavily abraded, and were soled at the store at half price.'} In this section,  we first describe the three components to generate homophonic puns as shown in Figure \ref{fig:illustration}: a context word and phrase selector, a label predictor and the procedure of curating training signals, and the generation module in Section \ref{subsec:obtain} to \ref{subsec:generation-module}. Then, we migrate the whole system to homographic puns in Section \ref{subsec:migrate}.
 
\subsection{Obtaining Context Words and Phrases}\label{subsec:obtain}
We retrieve and select two things: a context word that supports the meaning of the pun word, and a phrase that is both characteristic to the alternative word and
compatible with the pun word.

Inspired by~\citet{he2019pun}, given a pun-alternative word pair ($pw-aw$), we look for an ideal phrase that contains $aw$ and replace it with $pw$ to arouse surprise. To this end, we first extract multiple ($N_1$=20) phrases that contain $aw$ from a large non-pun corpus consisting of 20,000 sentences from Wikipedia and Gutenberg BookCorpus \cite{lebert2009short}, and rank the phrases by how well they exhibit the semantics of the pun pair. Specifically, we first replace $aw$ with a `<mask>' token, and run RoBERTa-Large \cite{liu2019roberta} to obtain the probability of $aw$ in the masked position. We remove the less probable half, filtering out those that are less characteristic of $aw$, as shown in Table \ref{table:rank}. Next, we conduct a similar mask infilling procedure for $pw$, and select the middle-ranking phrase to avoid it being either too general (e.g., `a new taxi was created') or too incompatible (e.g., `an export taxi on agricultural products'). These two rankings ensure the final selected phrase arouses surprisal when people see $pw$ instead of $aw$, but also still find it reasonable.

\begin{table}[!t]\vspace{-2mm}
\small
\centering
\begin{tabular}{@{}lll@{}}
\toprule
Retrieved Phrase                             & Tax &  Taxi \\ \midrule
get in all that <mask> trouble              & \checkmark   & \checkmark    \\
an export <mask> on agricultural products & \checkmark   & $\times$    \\
a new <mask> was created                    & \checkmark   & $\times$    \\
made <mask> deductible against income     & \checkmark   & \checkmark    \\ \bottomrule
\end{tabular}\vspace{-1.5mm}
\caption{\footnotesize An example of the retrieved phrases which are characteristic of the alternative word, `tax'. The pun-pair is `tax-taxi'. A `\checkmark' means the phrase is compatible with the corresponding word according to our mask in-filling model.}
\label{table:rank}
\end{table}

For obtaining the context words, our idea is similar to that proposed by \cite{mittal2022ambipun}. We retrieve sentences from the same non-pun corpus containing the target word ($pw$), and then extract keywords from the sentences using RAKE \cite{rose2010automatic}. Based on the TF-IDF values of those keywords, we take the top $N_2$ ($N_2$=20) words that uniquely co-occur with the target word, and then randomly sample one to encourage creativity.

\subsection{Label Predictor} \label{subsec:label_predictor}
\begin{table}[!t]\vspace{-2mm}
\small
\centering
\begin{tabular}{@{}llll@{}}
\toprule
Category                & A    & D1   & D2   \\ \midrule
$\mathbf{Bert_n}$         & 0.81 & 0.68 & 0.60  \\
-  human labeled train data    & - 0.02 & - 0.06 & - 0.05 \\
+ high confidence ($T$=0.9) & +0.03 & +0.02 & +0.05 \\ \bottomrule
\end{tabular}
\caption{\footnotesize The F1 scores of $\mathbf{Bert_n}$ and its ablations on human annotated testset.}
\label{table:breakdown}
\vspace{-1.5em}
\end{table}
Our label predictor aims at learning and predicting the word-level structures of 1,500 human-written pun sentences collected by \citet{miller2017semeval}. Each word in a pun sentence falls into one of three types: `A' for ambiguous, `D1' for distinct to the pun word, `D2' for distinct to the alternative word. We finetune a BERT  \cite{devlin2018bert} sequence classification model to \textit{predict} the \textit{next} token type.  In this section, we will first talk about a data-efficient label collection procedure that boosts from a weak, unsupervised method to a stronger, weakly supervised method . 
\paragraph{Ground Truth Label Curation} Before we could train the model, how can we automatically categorize each word in a pun sentence? We start with an unsupervised approach: word semantic similarity. Specifically, we compute the cosine similarity between the glove embeddings of $pw$ and $aw$ and each word in the sentence ($tw$), and label the word as D1/D2 if the difference is larger than a threshold $T$ (i.e., $|cos(tw, pw) - cos (tw, aw)| > T$). Otherwise, we label the word as A. We also compute their correlation with human judgements. In total, we collected human annotations on 500 data points. Since the label predictor should predict the type/category of the \textit{next} word without knowing the future, we mimic this setting and show human annotators 1) an incomplete pun sentence, i.e. containing the part of the sentence \textit{before} the current word $tw$ being evaluated, 2) $tw$, and 3) $pw$ and $aw$, and ask them to decide whether $tw$ is distinct to $pw$, distinct to $aw$, or ambiguous. With grid search, we find out that with optimal $T$ set to 0.15, the aforementioned purely unsupervised method gets 72.9\% labeling accuracy.\footnote{We have also tried using BERT embedding for semantic similarity. Yet the correlation with human judgment is much lower than the glove embeddings.}

To further improve the reliability of the `ground truth' labels, we finetune a BERT-base model as a sequence classifier to classify each word into the three categories. The intuition is to provide this BERT classifier with \textit{less noisy data} so that it can learn the task better than the unsupervised approach. The training data of this BERT classifier includes 8,000 automatic labels obtained using the word unsupervised method. A word is considered distinct (D1/D2) if the difference is > 1.5\textit{T}, and ambiguous if the difference is < \textit{T}. In order to compose a dataset with cleaner labels, we simply disregard those training samples whose semantic difference is [\textit{T}, 1.5\textit{T}].
We include the incomplete pun sentence, the current word, and the pun pairs as input. Using this, we are able to improve the `gold' label accuracy to 84.6\% on a human annotated held-out dataset. We call this model $\mathbf{Bert_c}$.

\paragraph{Training} The label predictor used in our framework predicts the type of the \textit{next} word that is yet to be generated. We call this model $\mathbf{Bert_n}$.\footnote{Note that there are two models: a classifier $\mathbf{Bert_c}$ and a predictor $\mathbf{Bert_n}$. The former knows the current token while the later does not. We need $\mathbf{Bert_n}$ in inference, and $\mathbf{Bert_c}$ is introduced to help curate $\mathbf{Bert_n}$'s training signals.} The training data of $\mathbf{Bert_n}$ comes from two parts: an additional 430 human labeled data that the unsupervised method and the $\mathbf{Bert_c}$ disagree on, and 8,000 automatic labels where both models agree. A breakdown of its performance by category is reported in Table \ref{table:breakdown}. In addition, we could further improve the predictor's F1 score by considering its confidence. We gain an average of 14.9\% increase by discarding only 9.8\% cases that the label predictor is less confident on. We argue that there can be multiple choices for the next word, and hence the best performance of this task is bounded. Considering that the task of predicting the type of the next word is much harder because there can be multiple choices as next words, we take into account the confidence level when we use it in the next step. 

\begin{algorithm}[!t]
   \caption{\footnotesize Discriminative Generation (Single Step)}
   \small
    \begin{algorithmic}[1]

      \Function{DiscriminativeGen}{}\\
      \textbf{Parameters: }{pun word (pw), alter word (aw), predicted next word label $L$, confidence $c$} and its threshold $T$\\
      $cands$ = gpt2.gen\_next\_word(num = N) 
        \If{$c > T$} 
        \If {$L$==A} sort($cands$, pw+aw)  \EndIf
        \If {$L$==D1} sort($cands$, pw) \EndIf 
        \If {$L$==D2} sort($cands$, aw) \EndIf 
        \EndIf
        \Comment{Sort according to semantic relevance in Section \ref{subsec:label_predictor}}
        \Return{$cands$}
       \EndFunction
\end{algorithmic}
\label{algo:1}
\end{algorithm}\vspace*{-1mm}

\subsection{Generation Module} \label{subsec:generation-module}
\paragraph{Data Preparation and Fine-tuning} We finetune the GPT-2 model on a combination of Gutenberg BookCorpus and jokes \cite{annamoradnejad2020colbert} to learn the \textit{task format}: given a keyword and a phrase as input, generate a sentence containing them both. For each sentence in the corpus, we use RAKE to extract two salient words. We then include the surrounding context around one of the salient words as the phrase. We make sure that positions of the two extracted keywords are far away enough that the phrase does not contain the other extracted word.
\paragraph{Inference}
At inference time, we feed the phrase and context word obtained in Section \ref{subsec:obtain} as input, and steer the finetuned language model using the label predictor. At each step, we get the predicted type of next token, $L$, with the corresponding confidence $c$. If $c$ is larger than a threshold $T$, we score and rerank the candidate next words  by the corresponding label, which can be found in Algorithm \ref{algo:1}. Otherwise, if our label predictor is less confident, we do not intervene in the language model's generation. We also enforce the appearance of the complete phrase during decoding when the first two words in the phrase have been generated.

\subsection{Migrate to Homographic Puns}\label{subsec:migrate}
We convert the task of generating homographic puns to homophonic puns by leveraging a word-sense disambiguation (WSD) model \cite{bevilacqua-navigli-2020-breaking}. For example, if the target pun word is ``sentence" and the two sense definitions are \textit{``a set of words..."} and \textit{``the punishment..."}, we run the WSD model to identify which extracted phrases exhibit the second sense. Next, we obtain two new words using a reverse dictionary \cite{qi2020wantwords}: \textit{`clause'} for the first sense and \textit{`conviction'} for the second. Then the task can be viewed as that for homophonic puns, where the substitute $pw$ is `clause' and the substitute $aw$ is `conviction'. The rest of the generation process is the same as in Section \ref{subsec:label_predictor} and \ref{subsec:generation-module}.

\section{Experiments}
\subsection{Compared Models}\label{subsec: baselines}
We compare with the best two existing models for each pun type. \textbf{Homophonic}: SurGen \cite{he2019pun}, a retreive-and-edit model using the local-global surprisal principle; and LCR \cite{yu-etal-2020-homophonic}, the SOTA model that first finds appropriate lexical constraints and then rewrites the sentence. \textbf{Homographic}: Pun-GAN \cite{luo2019pun}, a model that adopts GANs to encourage ambiguity; AmbiPun \cite{mittal2022ambipun}, the SOTA model that generates puns by including contexts words from both senses. We also compare the ablations of our own models: the base GPT-2 model where a random word and phrase is given, the base model with the label predictor added or the selected context word and phrase (which we call `+ select') added, and best model that includes both the label predictor and selection.

\begin{table}[!t]\vspace{-2mm}
\small
\centering
\begin{tabular}{@{\ \ \ }l@{\ \ \ }c@{\ \ \ }c@{\ \ \ }c@{\ \ \ }c@{\ \ \ }c@{\ \ \ }}
\toprule
\rowcolor[HTML]{EFEFEF}
\textbf{Homophonic Pun}                  & A              & D1             & D2            & Avg D     & S             \\ \midrule
LCR               & \textbf{24.93} & 4.57           & 0.90          & 2.73          & 0.15          \\
SurGen           & {\ul 23.18}    & 1.65           & 3.94          & 2.80          & 0.54          \\ \midrule
Base GPT-2    & 14.85          & 0.77           & 0.72          & 0.74          & -0.54         \\
+ label predictor$^*$ & 18.90          & 3.83           & 2.69          & 3.26          & 0.53          \\
+ select$^*$         & 17.56          & 3.72           & 4.22          & 3.97          & 0.68          \\
+ both$^*$           & 20.27          & {\ul 6.71}     & {\ul 5.46}    & {\ul 6.09}    & \textbf{0.77} \\ \midrule
Human             & 22.50          & \textbf{11.72} & \textbf{6.17} & \textbf{8.95} & {\ul 0.71}    \\ \bottomrule
\end{tabular}

\begin{tabular}{@{ }l@{\ \ \ }c@{\ \ }c@{\ \ \ }c@{\ \ \ }c@{\ \ \ }c@{ }}
\toprule
\rowcolor[HTML]{EFEFEF} \textbf{Homographic Pun}                  & A              & D1             & D2            & Avg D     & S             \\ \midrule
Pun-GAN            &\multicolumn{1}{c}{{\ul 20.12}} & 1.22                     & 1.08                     & 1.15       & 0.22       \\
AmbiPun &\multicolumn{1}{c}{17.14} &2.39          &2.01          &2.20          &0.34          \\ \midrule
Base GPT-2    & 18.45                           & 0.84                     & 0.82                     & 0.83       & -0.21      \\
+ label predictor$^*$ &\textbf{20.64}                  & 4.54                     & 3.18                     & 3.86       & 0.45       \\
+ select$^*$         & {\ul 19.80}                     &2.89 &3.20 & 3.05       & 0.59       \\
+ both$^*$            &{18.27}       &{\ul 5.87}               &{\ul 6.62}               &{\ul 6.25} &{\ul 0.80} \\  \midrule
Human   & 18.00                     & {\textbf{7.01}} & {\textbf{8.76}} & \textbf{7.88} & \textbf{0.83} \\ \bottomrule
\end{tabular}
\vspace{-2mm}
\caption{\footnotesize Results of automatic evaluation on ambiguity (A), distinctiveness to pun word (D1) and alternative word (D2), and surprisal ratio (S). The $^*$ indicates ablations of our method where paired t-test shows that the difference between our best performing model and the best baseline is statistically significant (p<0.05). Boldface denotes the best score and underline denotes the second best. }
\label{table:auto-eval-phonic}
\vspace{-5mm}
\end{table}

\subsection{Automatic Evaluation}
For each system, we compute the ambiguity, distinctiveness, and the surprisal ratio \cite{kao2016computational, he2019pun}, and report the results in Table \ref{table:auto-eval-phonic}. For both pun types, our model surpasses the best baseline by a large margin in terms of distinctiveness, meaning that our model supports distinct viewpoints in the sentence. Notably, our surprisal ratio surpasses that in human-written puns.

Moreover, \citet{he2019pun} have shown that while higher D and S scores usually indicate higher quality, that is not the case for ambiguity. Intuitively, since many ambiguous sentences are not informative (e.g. “I went to the bank”), ambiguity alone is insufficient. Our results correlate with the findings that A is useful in distinguishing non-puns from puns, while D and S are useful when spotting good and funny puns from bad or boring puns. Besides, our statistics show that human tend to context the pun word more when writing homophonic puns: 24\% of the words are distinct to the pun word, versus 14\% for the alternative word. This partially explains the imbalance between D1 and D2 for human-written puns. Our label predictor also learns such distribution and steers base GPT-2 more towards the pun word.

\begin{table}[t]\vspace{-2mm}
\small
\centering
\begin{tabular}{@{}llll@{}}
\toprule
\rowcolor[HTML]{EFEFEF}
\textbf{Homophonic Pun}     & Success       & Informative      & Funny         \\\midrule
LCR           & 39\%          & 2.11          & 2.14          \\
SurGen        & 42\%          & 2.74          & 2.35          \\ \midrule
Base GPT-2  & 16\%          & 2.52          & 1.56          \\
+ label predictor$^*$      & 40\%          & 2.96          & 2.04          \\
+ select$^*$ & 49\%          & 3.35          & 2.57          \\
+ both$^*$    & {\ul 56\%}          & {\ul 3.60}    & {\ul 2.96}    \\ \midrule
Human         & \textbf{89\%} & \textbf{4.56} & \textbf{4.04} \\ \bottomrule
\end{tabular}
\quad
\begin{tabular}{@{}llll@{}}
\toprule
\rowcolor[HTML]{EFEFEF} \textbf{Homographic Pun}           & Success       & Informative      & Funny         \\\midrule
Pun-GAN            & 20\%          & 1.72          & 1.54          \\
AmbiPun           & 44\%          & 2.76          & 2.40          \\ \midrule
Base GPT-2         & 14\%          & 2.17          & 1.55          \\
+ label predictor$^*$ & 29\%          & 2.84          & 2.03          \\
+ select$^*$         & 43\%          & 3.13          & 2.51          \\
+ both$^*$            & {\ul 47\%}    & {\ul 3.32}    & {\ul 2.83}    \\ \midrule
Human             & \textbf{85\%} & \textbf{4.23} & \textbf{3.87} \\ \bottomrule
\end{tabular}
\vspace{-2mm}
\caption{Results of human evaluation on pun success rate, informativeness and funniness. The $^*$ indicates ablations of our method. Boldface denotes the best score and underline denotes the second best.
}
\label{table:human-eval}
 \vspace{-3.7 mm}
\end{table}

\subsection{Human Evaluation}
We ask qualified workers to judge if the given sentence is a successful pun, and rate the informativeness (or specificness) and funniness on a scale from 1 to 5. The evaluation details can be found in Appendix \ref{sec:appendix-eval}.
Results in Table~\ref{table:human-eval} show that our model achieves the highest success rate and is the most informative and funny among all machines. We also observe that the improvements over homographic puns are smaller than that of homophonic puns. Upon error analysis, we find that half of the failure cases of homographic puns are due to inappropriate substitute pun/alternative words. Instead of using the sense keys provided in WordNet \cite{miller1995wordnet}, if the user can manually provide the sense definitions to ensure the substitute pun pair is reasonable, such a bottleneck shall be resolved. 

\subsection{Ablation and Case Study}
To validate the effectiveness of each proposed module, we report their performance in Table~\ref{table:auto-eval-phonic} and \ref{table:human-eval} and a bar chart in Appendix \ref{sec:appendix-eval} for more straightforward illustration. Both the label predictor and the word/phrase selection process positively contribute to the outputs, and it works best when combined.

\definecolor{mypurple}{HTML}{6200C9}
\definecolor{myblue}{HTML}{0432FF}

\begin{table}[t]
\small
\begin{tabular}{@{}lp{19.5em}@{}} 
\toprule \vspace{-1mm}Pun pair & \textcolor{mypurple}{mane}-\textcolor{myblue}{main} \\ \midrule 
LCR        & The   mane object of the hair was accomplished.                                                                           \\
SurGen    & \begin{tabular}[c]{@{}p{20em}@{}}A   trot later, he was sitting away from the mane\\dining area.\end{tabular}      \\
Ours       & \begin{tabular}[c]{@{}p{20em}@{}}In some places, hair also makes up the mane \\entrance to fashion salons.\end{tabular} \\
Human & \begin{tabular}[c]{@{}p{19em}@{}}Lions don't have to worry about every little \\detail in life, just the mane thing.\end{tabular} \\ \bottomrule\vspace{-1mm}
\end{tabular}
\vspace{-2mm}
\begin{tabular}{@{}lp{19em}@{}}
\toprule \vspace{-1mm}Pun pair & sentence $\Longrightarrow${\color[HTML]{6200C9}clause}-{\color[HTML]{0432FF}punishment}                    \\ \midrule
Pun-GAN   & Due to the sentence it is in the United States. \\
AmbiPun & \begin{tabular}[c]{@{}p{19em}@{}}The sentence is ungrammatical. The jury didn't\\hear it.\end{tabular}              \\
Ours    & \begin{tabular}[c]{@{}p{20em}@{}}The language on a two-page sentence for fraud\\is full of guilt.\end{tabular}      \\
Human   & \begin{tabular}[c]{@{}p{20em}@{}}The judge has got a stutter. Looks like I am\\not getting a sentence.\end{tabular} \\ \bottomrule\vspace{-1mm}
\end{tabular}
\vspace{-3 mm}
\caption{Example outputs of different models. The pun pairs are randomly selected.}
\label{table:case-study}
 \vspace{-5.7 mm}
\end{table}

A comparison between our model and the baselines is in Table \ref{table:case-study}. Although existing approaches also include related words for semantic relevance, they tend to be too vague (e.g. LCR and Pun-GAN) or abrupt (e.g. `a trot later' by SurGen). We also showcase the outputs for two more pun pairs 
along with the predicted token types in Appendix \ref{appendix:more-example}. Both results demonstrate that our model is best at generating informative puns with humorous effects.

\section{Conclusion}

We propose a novel pun generation approach that incorporates three humor principles. To this end, we learn the sentence structures from human-written puns, and convert the task of homographic pun generation to homophonic pun generation. Our model achieves strong performance for both types. 

\section{Limitations}
We discuss several limitations of this work to inspire future research directions. First, our method rely on a small amount of human written puns as a training corpora, and thus might not work well for low resource languages. Second, as can be seen in Table \ref{table:breakdown}, the overall performance of the label predictor is not perfect. While we argue that the task of predicting the type of the next word is naturally difficult as there can be multiple good candidates, the errors of the label predictor may propagate and lead to unnatural outputs. 

Third, our system could not generate homographic puns as successfully as homophonic ones. Human evaluation and further error analysis show that the main reason of failure is that the generated substitute pun-alternative word pair is bad. Given a homographic pun word, we are currently retrieving its two sense definitions from WordNet \cite{miller1995wordnet} using the sense keys provided in the SemEval 2017 dataset \cite{miller2017semeval}, where the retrieved sense definitions are sometimes imprecise. Future directions include refining the procedure of finding the substitute pun-alternative word pairs, and curating a more accurate definition dataset for homograpic pun words.

Our proposed method is independent of the specific language model being used. The selection process is purely unsupervised and our label predictor can be theoretically combined with any language model as long as we can obtain the top k tokens it produces. Another future direction includes applying our technique to steer the GPT-3 \cite{brown2020language} or GPT-J model\footnote{https://huggingface.co/EleutherAI/gpt-j-6B} to generate humorous puns. 

\section*{Acknowledgments}
The authors would like to thank the members of PLUSLab and the anonymous reviewers for helpful comments. The research is suported in part by a Meta SRA. Yufei Tian is supported by an Amazon Fellowship.

\section*{Ethics}
Our proposed methods are based on the pretrained language model. It is known that pretrained language models could capture the bias reflected in the training data \cite{sheng2019woman,wallace2019universal}. Considering the nature of humorous puns, our models may potentially generate offensive content for certain groups or individuals.  We suggest to carefully examine the potential biases before deploying the models to real-world applications.

\bibliography{anthology,custom}

\begin{thebibliography}{23}
\expandafter\ifx\csname natexlab\endcsname\relax\def\natexlab#1{#1}\fi

\bibitem[{Annamoradnejad and Zoghi(2020)}]{annamoradnejad2020colbert}
Issa Annamoradnejad and Gohar Zoghi. 2020.
\newblock Colbert: Using bert sentence embedding for humor detection.
\newblock \emph{arXiv preprint arXiv:2004.12765}.

\bibitem[{Bevilacqua and Navigli(2020)}]{bevilacqua-navigli-2020-breaking}
Michele Bevilacqua and Roberto Navigli. 2020.
\newblock \href {https://doi.org/10.18653/v1/2020.acl-main.255} {Breaking
  through the 80{\%} glass ceiling: {R}aising the state of the art in word
  sense disambiguation by incorporating knowledge graph information}.
\newblock In \emph{Proceedings of the 58th Annual Meeting of the Association
  for Computational Linguistics}, pages 2854--2864, Online. Association for
  Computational Linguistics.

\bibitem[{Braslavski et~al.(2018)Braslavski, Blinov, Bolotova, and
  Pertsova}]{braslavski2018evaluate}
Pavel Braslavski, Vladislav Blinov, Valeria Bolotova, and Katya Pertsova. 2018.
\newblock How to evaluate humorous response generation, seriously?
\newblock In \emph{Proceedings of the 2018 Conference on Human Information
  Interaction \& Retrieval}, pages 225--228.

\bibitem[{Brown et~al.(2020)Brown, Mann, Ryder, Subbiah, Kaplan, Dhariwal,
  Neelakantan, Shyam, Sastry, Askell et~al.}]{brown2020language}
Tom~B Brown, Benjamin Mann, Nick Ryder, Melanie Subbiah, Jared Kaplan, Prafulla
  Dhariwal, Arvind Neelakantan, Pranav Shyam, Girish Sastry, Amanda Askell,
  et~al. 2020.
\newblock Language models are few-shot learners.
\newblock \emph{arXiv preprint arXiv:2005.14165}.

\bibitem[{Devlin et~al.(2018)Devlin, Chang, Lee, and
  Toutanova}]{devlin2018bert}
Jacob Devlin, Ming-Wei Chang, Kenton Lee, and Kristina Toutanova. 2018.
\newblock Bert: Pre-training of deep bidirectional transformers for language
  understanding.
\newblock \emph{arXiv preprint arXiv:1810.04805}.

\bibitem[{Hashimoto et~al.(2018)Hashimoto, Guu, Oren, and
  Liang}]{hashimoto2018retrieveandedit}
Tatsunori~B. Hashimoto, Kelvin Guu, Yonatan Oren, and Percy Liang. 2018.
\newblock \href {http://arxiv.org/abs/1812.01194} {A retrieve-and-edit
  framework for predicting structured outputs}.

\bibitem[{He et~al.(2019)He, Peng, and Liang}]{he2019pun}
He~He, Nanyun Peng, and Percy Liang. 2019.
\newblock Pun generation with surprise.
\newblock In \emph{2019 Conference of the North American Chapter of the
  Association for Computational Linguistics: Human Language Technologies, NAACL
  HLT 2019}, pages 1734--1744. Association for Computational Linguistics (ACL).

\bibitem[{Kao et~al.(2016)Kao, Levy, and Goodman}]{kao2016computational}
Justine~T Kao, Roger Levy, and Noah~D Goodman. 2016.
\newblock A computational model of linguistic humor in puns.
\newblock \emph{Cognitive science}, 40(5):1270--1285.

\bibitem[{Lebert(2009)}]{lebert2009short}
Marie Lebert. 2009.
\newblock A short history of ebooks.

\bibitem[{Liu et~al.(2019)Liu, Ott, Goyal, Du, Joshi, Chen, Levy, Lewis,
  Zettlemoyer, and Stoyanov}]{liu2019roberta}
Yinhan Liu, Myle Ott, Naman Goyal, Jingfei Du, Mandar Joshi, Danqi Chen, Omer
  Levy, Mike Lewis, Luke Zettlemoyer, and Veselin Stoyanov. 2019.
\newblock Roberta: A robustly optimized bert pretraining approach.
\newblock \emph{arXiv preprint arXiv:1907.11692}.

\bibitem[{Luo et~al.(2019)Luo, Li, Yang, Li, Chang, Sui, and Sun}]{luo2019pun}
Fuli Luo, Shunyao Li, Pengcheng Yang, Lei Li, Baobao Chang, Zhifang Sui, and
  Xu~Sun. 2019.
\newblock Pun-gan: Generative adversarial network for pun generation.
\newblock In \emph{Proceedings of the 2019 Conference on Empirical Methods in
  Natural Language Processing and the 9th International Joint Conference on
  Natural Language Processing (EMNLP-IJCNLP)}, pages 3388--3393.

\bibitem[{Miller(1995)}]{miller1995wordnet}
George~A Miller. 1995.
\newblock Wordnet: a lexical database for english.
\newblock \emph{Communications of the ACM}, 38(11):39--41.

\bibitem[{Miller et~al.(2017)Miller, Hempelmann, and
  Gurevych}]{miller2017semeval}
Tristan Miller, Christian~F Hempelmann, and Iryna Gurevych. 2017.
\newblock Semeval-2017 task 7: Detection and interpretation of english puns.
\newblock In \emph{Proceedings of the 11th International Workshop on Semantic
  Evaluation (SemEval-2017)}, pages 58--68.

\bibitem[{Mittal et~al.(2022)Mittal, Tian, and Peng}]{mittal2022ambipun}
Anirudh Mittal, Yufei Tian, and Nanyun Peng. 2022.
\newblock Ambipun: Generating humorous puns with ambiguous context.
\newblock \emph{NAACL}.

\bibitem[{Qi et~al.(2020)Qi, Zhang, Yang, Liu, and Sun}]{qi2020wantwords}
Fanchao Qi, Lei Zhang, Yanhui Yang, Zhiyuan Liu, and Maosong Sun. 2020.
\newblock Wantwords: An open-source online reverse dictionary system.
\newblock In \emph{Proceedings of the 2020 Conference on Empirical Methods in
  Natural Language Processing: System Demonstrations}, pages 175--181.

\bibitem[{Rose et~al.(2010)Rose, Engel, Cramer, and Cowley}]{rose2010automatic}
Stuart Rose, Dave Engel, Nick Cramer, and Wendy Cowley. 2010.
\newblock Automatic keyword extraction from individual documents.
\newblock \emph{Text mining: applications and theory}, 1:1--20.

\bibitem[{Sheng et~al.(2019)Sheng, Chang, Natarajan, and Peng}]{sheng2019woman}
Emily Sheng, Kai-Wei Chang, Premkumar Natarajan, and Nanyun Peng. 2019.
\newblock The woman worked as a babysitter: On biases in language generation.
\newblock \emph{arXiv preprint arXiv:1909.01326}.

\bibitem[{Sun et~al.(2022{\natexlab{a}})Sun, Narayan-Chen, Oraby, Cervone,
  Chung, Huang, Liu, and Peng}]{sun2022expun}
Jiao Sun, Anjali Narayan-Chen, Shereen Oraby, Alessandra Cervone, Tagyoung
  Chung, Jing Huang, Yang Liu, and Nanyun Peng. 2022{\natexlab{a}}.
\newblock Expunations: Augmenting puns with keywords and explanations.
\newblock In \emph{Proceedings of the 2022 Conference on Empirical Methods in
  Natural Language Processing (EMNLP)}.

\bibitem[{Sun et~al.(2022{\natexlab{b}})Sun, Narayan-Chen, Oraby, Gao, Chung,
  Huang, Liu, and Peng}]{sun2022contextpun}
Jiao Sun, Anjali Narayan-Chen, Shereen Oraby, Shuyang Gao, Tagyoung Chung, Jing
  Huang, Yang Liu, and Nanyun Peng. 2022{\natexlab{b}}.
\newblock Context-situated pun generation.
\newblock In \emph{Proceedings of the 2022 Conference on Empirical Methods in
  Natural Language Processing (EMNLP)}.

\bibitem[{Wallace et~al.(2019)Wallace, Feng, Kandpal, Gardner, and
  Singh}]{wallace2019universal}
Eric Wallace, Shi Feng, Nikhil Kandpal, Matt Gardner, and Sameer Singh. 2019.
\newblock Universal adversarial triggers for attacking and analyzing nlp.
\newblock \emph{arXiv preprint arXiv:1908.07125}.

\bibitem[{Yu et~al.(2018)Yu, Tan, and Wan}]{yu2018neural}
Zhiwei Yu, Jiwei Tan, and Xiaojun Wan. 2018.
\newblock A neural approach to pun generation.
\newblock In \emph{Proceedings of the 56th Annual Meeting of the Association
  for Computational Linguistics (Volume 1: Long Papers)}, pages 1650--1660.

\bibitem[{Yu and Wan(2019)}]{yu-wan-2019-avoid}
Zhiwei Yu and Xiaojun Wan. 2019.
\newblock \href {https://doi.org/10.18653/v1/N19-1092} {How to avoid sentences
  spelling boring? towards a neural approach to unsupervised metaphor
  generation}.
\newblock In \emph{Proceedings of the 2019 Conference of the North {A}merican
  Chapter of the Association for Computational Linguistics: Human Language
  Technologies, Volume 1 (Long and Short Papers)}, pages 861--871, Minneapolis,
  Minnesota. Association for Computational Linguistics.

\bibitem[{Yu et~al.(2020)Yu, Zang, and Wan}]{yu-etal-2020-homophonic}
Zhiwei Yu, Hongyu Zang, and Xiaojun Wan. 2020.
\newblock \href {https://doi.org/10.18653/v1/2020.emnlp-main.229} {Homophonic
  pun generation with lexically constrained rewriting}.
\newblock In \emph{Proceedings of the 2020 Conference on Empirical Methods in
  Natural Language Processing (EMNLP)}, pages 2870--2876, Online. Association
  for Computational Linguistics.

\end{thebibliography}
\bibliographystyle{acl_natbib}

\cleardoublepage
\appendix

\section*{Appendix}

\section{Implementation Details}\label{subsec:implement}
\subsection{Experimental Settings}
 For the label predictor described in section \ref{subsec:label_predictor}, we finetune the bert-base model for 6 epochs. The training time is only 20 minutes on a single NVIDIA A100 GPU. As for the finetuning the language model to learn the task of generating a sentence that includes the given keyword and phrase described in Section \ref{subsec:generation-module}, we finetune GPT-2 large for 6 epochs and the training time is about 3 hours on a single NVIDIA A100 GPU. The max sequence length for target and source is set to 50 and batch size is set to 6. 

\subsection{Pretrained Model Selection} 
For sequence classification tasks -- improving the reliability of the `ground truth' labels that contribute towards training data for the label predictor and the label prediction task itself, wee compare the base and large variants of the BERT and RoBERTa models and find no significant improvement using RoBERTa or the large variants, which is why we stick with the smallest BERT-base model. For the mask-infilling task towards obtaining an ideal phrase, we do find an improvement using RoBERTa-Large over other variants.

\subsection{Human Annotation to Train/Test the Label Predictor} 
Recall that we collected human annotation on 430 word pairs to boost the performance of $\mathbf{Bert_c}$ and another 500 word pairs to test the model. Each data point is annotated by three people. Namely, given $tw$, $pw$, and $aw$, the annotators are asked to decide whether $tw$ is distinct to $pw$, $aw$, or ambiguous. The inter annotator agreement is 0.62 for Krippendorff's alpha, indicating a strong agreement.

\subsection{Decoding Details}
Since GPT-2 uses BPE tokenizer and generate a subword at each decoding step, we ask the finetuned GPT-2 model to continue generating subwords until a complete word is generated. In addition, we keep a beam size of 20 at all time, so that we could obtain a list of complete words as candidate next words.

\section{Evaluation}\label{sec:appendix-eval}
\begin{figure}[t!]
\footnotesize
  \centering
    \includegraphics[width=0.48\textwidth]{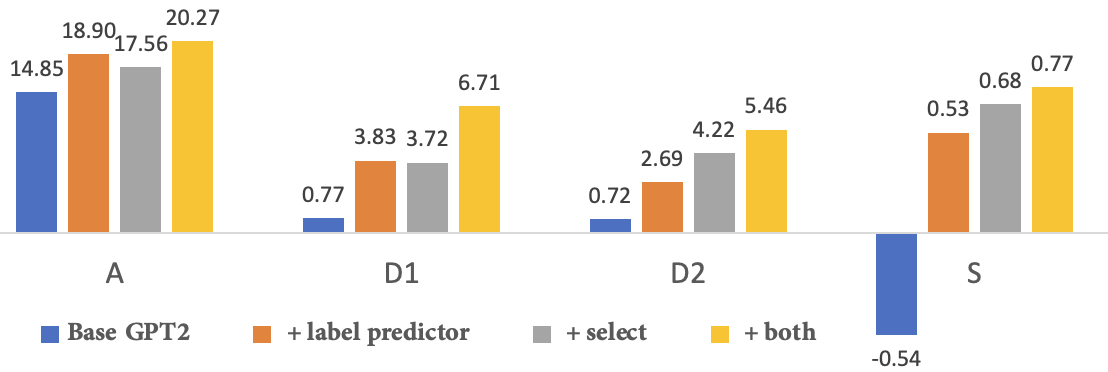}
  \caption{Bar chart showing the improvement after adding each component on homophonic puns.}
  \label{fig:bar-chart}    
  \vspace{-3.7 mm}
\end{figure}

\paragraph{Implementation of the Ambiguity-Distinctiveness Model} The initial probabilistic model of ambiguity and distinctiveness is proposed by \citet{kao2016computational}, yet \citet{he2019pun} improve the implementation by training a skip-gram model to avoid human labor. For more straightforward illustration, the bar chart in Figure \ref{fig:bar-chart} shows the improvement after adding each component on homophonic puns. Note that the numbers in this bar chart is the same as that in Table \ref{table:auto-eval-phonic} in the main paper.   

\paragraph{Human Evaluation}The workers are paid \$20 per hour. For pun success judgement (yes/no question), we take the majority vote among three workers, while for specificness and funniness (1 to 5), we take the average ratings. We then use the pairwise kappa coefficient to measure the inter-annotator agreement (IAA). The average inter-annotator agreement of all raters for pun success, specificness and funniness are 0.59, 0.44 and 0.47, meaning that annotators moderately agree with each other. Considering the subjectivity of this task \cite{braslavski2018evaluate}, we argue that our collected results are reasonably reliable for the purpose of pun generation. Besides, we conducted paired t-test and show that the success rate and funniness ratings of our systems differentiate from the best baseline model with statistical significance (p-value < 0.05).

\begin{figure}[t!]
\footnotesize
  \centering
    \includegraphics[width=0.4\textwidth]{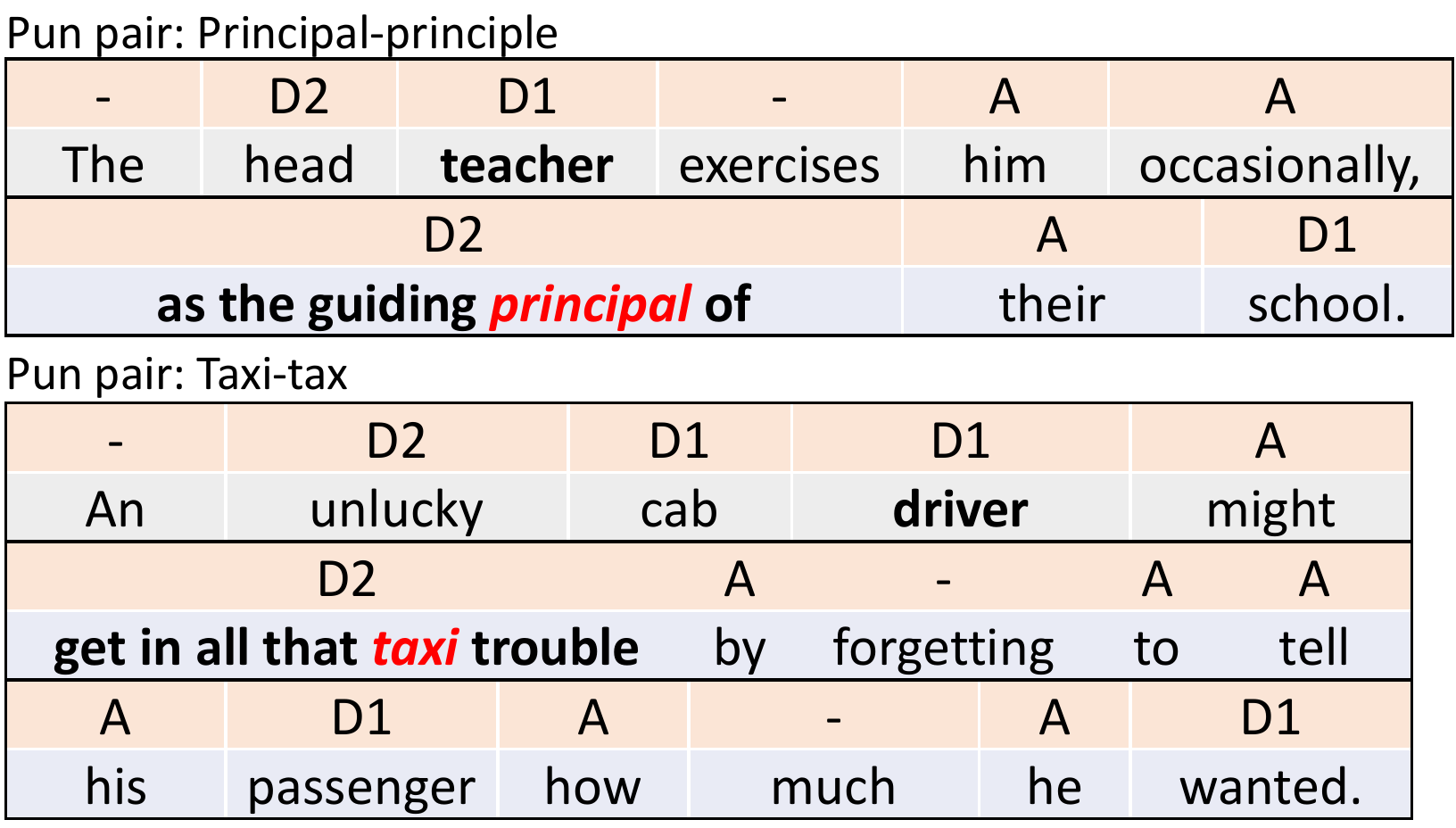}
    \vspace{-1.7 mm}
  \caption{Two randomly sampled examples generated by our model and the predicted labels. The context words and the extracted phrases are in boldface.}
  \label{fig:more_examples}    
  \vspace{-3.7 mm}
\end{figure}

\section{More generated examples by our model}\label{appendix:more-example}
Figure \ref{fig:more_examples} shows two randomly sampled examples generated by our model and the predicted labels. We can see that the extracted phrases (i.e., `as the guiding principal of' and `get in all that taxi trouble') arouses surprise when people read it the local context, while making sense in the global context. In addition, the label predictor successfully steer the base GPT-2 model to generate more ambiguously or distinctively at each step, resulting in humorous puns.

\end{document}